\documentclass[11pt]{article}

\usepackage[utf8]{inputenc}
\usepackage[T1]{fontenc}
\usepackage{amsmath,amsfonts,amssymb}
\usepackage{graphicx}
\usepackage{subcaption}
\usepackage{hyperref}
\usepackage{cleveref}
\usepackage[margin=1in]{geometry}
\usepackage{natbib}
\usepackage{authblk}
\usepackage{xcolor}

\title{Computational Turing Test Reveals Systematic Differences Between Human and AI Language}

\author[1]{Nicolò Pagan\thanks{Corresponding author: nicolo.pagan@uzh.ch}}
\author[2]{Petter Törnberg}
\author[3]{Christopher A. Bail}
\author[1]{Anikó Hannák}
\author[4]{Christopher Barrie}

\affil[1]{University of Zurich, Department of Informatics, Zurich, Switzerland}
\affil[2]{University of Amsterdam, Institute for Logic, Language and Computation (ILLC), Amsterdam, The Netherlands}
\affil[3]{Duke University, Durham, NC, USA}
\affil[4]{New York University, Department of Sociology, New York City, USA}

\date{}

\begin{document}

\maketitle

\begin{abstract}
Large language models (LLMs) are increasingly used in the social sciences to simulate human behavior, based on the assumption that they can generate realistic, human-like text. Yet this assumption remains largely untested. Existing validation efforts rely heavily on human-judgment-based evaluations -- testing whether humans can distinguish AI from human output -- despite evidence that such judgments are blunt and unreliable. As a result, the field lacks robust tools for assessing the realism of LLM-generated text or for calibrating models to real-world data. This paper makes two contributions. First, we introduce a \textit{computational Turing test}: a validation framework that integrates aggregate metrics (BERT-based detectability and semantic similarity) with interpretable linguistic features (stylistic markers and topical patterns) to assess how closely LLMs approximate human language within a given dataset. Second, we systematically compare nine open-weight LLMs across five calibration strategies -- including fine-tuning, stylistic prompting, and context retrieval -- benchmarking their ability to reproduce user interactions on X (formerly Twitter), Bluesky, and Reddit. Our findings challenge core assumptions in the literature. Even after calibration, LLM outputs remain clearly distinguishable from human text, particularly in affective tone and emotional expression. Instruction-tuned models underperform their base counterparts, and scaling up model size does not enhance human-likeness. Crucially, we identify a trade-off: optimizing for human-likeness often comes at the cost of semantic fidelity, and vice versa. These results provide a much-needed scalable framework for \textit{validation} and \textit{calibration} in LLM simulations -- and offer a cautionary note about their current limitations in capturing human communication.
\end{abstract}


\section{Introduction}

Large language models (LLMs) have rapidly become indispensable tools across the social sciences, powering tasks from text annotation and survey design to synthetic data generation \citep{tornberg2023simulating,gilardi2023chatgpt,ziems2024can}. Increasingly, researchers use LLMs not only as instruments of analysis but as proxies for human behavior. By simulating realistic dialogue and decision-making processes, LLMs promise to act as consistent and scalable stand-ins for human participants \citep{argyle2023leveraging,flamino2024limits}, opening new horizons for generative social simulations that go beyond traditional agent-based models \citep{park2023generative,guo2024large,liu2025mosaic}. These applications rest on a crucial assumption: that LLMs can convincingly reproduce human communication.

Yet this assumption has rarely been rigorously tested. Scholars have emphasized two unresolved challenges for LLM-based simulation \citep{larooij2025large,sen2025validating}: \textit{validation} -- determining whether model outputs are truly realistic -- and \textit{calibration} -- aligning models with the linguistic and emotional patterns of human communication \citep{bail2024can,anthis2025llm,grossmann2023ai}. Most evaluations still rely on human judgment, asking individuals whether text ``sounds'' human \citep{park2023generative}. 
{Recent work demonstrates that LLM-generated social media conversations can successfully pass such human-judgment-based Turing tests \citep{bouleimen2025collective}, yet this apparent success may reflect the limitations of human perception rather than genuine linguistic equivalence.} 
Such tests are subjective, unscalable, and set a low bar: humans routinely overlook subtle linguistic and emotional cues that distinguish authentic from synthetic text \citep{clark2021humannotgold,larooij2025large}. 
Similarly, most calibration efforts depend on ad hoc prompt engineering rather than systematic optimization or post-generation selection \citep{ng2025llm,cho2024llm}. Recent studies warn that these approaches can distort group-level features and reduce realism \citep{wang2024large,bisbee2024synthetic,santurkar2023whose}, and that participants in experiments often perceive LLM confederates as less convincing than real humans \citep{flamino2024limits}.

In this paper, we address both challenges. First, we develop a computational Turing test -- a scalable validation framework that assesses the realism of AI-generated language across three dimensions: (1) detectability, measuring how easily human and AI text can be distinguished; (2) semantic fidelity, quantifying similarity in meaning to human reference replies; and (3) interpretable linguistic analysis, identifying the stylistic and topical features that reveal AI authorship. Like the original Turing test, our framework evaluates the extent to which machine outputs are indistinguishable from human ones -- but does so by measuring linguistic similarity directly, rather than relying on human perception. This framework offers a systematic, reproducible, and more rigorous alternative to human evaluation.

Second, we conduct a \textit{comprehensive calibration benchmark}, evaluating how different optimization strategies affect human-likeness. We test nine open-weight LLMs under two levels of calibration. The first compares five prompt-engineering configurations -- baseline, persona descriptions, stylistic examples, context retrieval, and fine-tuning. The second applies post-generation optimization, generating 20 candidate replies and selecting the best either for semantic similarity (cosine-optimal) or for minimal detectability (ML-optimal).

We use our validation framework to benchmark the models' abilities to reproduce user interactions on X (formerly Twitter), Bluesky, and Reddit. Social media provides an ideal testbed: it is central to model training data and a key domain of applied social simulation \citep{tornberg2023simulating,orlando2025generativeagentbasedmodelingreplicate,de2023emergence,park2024generative,gao2023s3}.

Contrary to common assumptions, our findings indicate that even the most advanced models and calibration methods remain readily distinguishable from human text: the computational Turing test identifies AI-generated text with 70–80\% accuracy -- well above chance. While optimization can mitigate some structural differences, deeper emotional and affective cues persist as reliable discriminators. Some sophisticated strategies, such as fine-tuning and persona descriptions, fail to improve realism or even make text more detectable, whereas simple stylistic examples and context retrieval yield modest gains.

Most notably, we identify a fundamental trade-off between realism and meaning: optimizing for reduced detectability lowers semantic fidelity, while optimizing for semantic accuracy does not consistently improve human-likeness. Together, these findings show that stylistic human-likeness and semantic accuracy are competing rather than aligned objectives—revealing persistent linguistic gaps between human and machine communication and setting a new benchmark for validating and calibrating language-based social simulations.

\section{Data, Models, \& Metrics}

\subsection{Data}
Our dataset comprises social media conversations collected from three platforms: Twitter/X, Bluesky, and Reddit. For Twitter/X, we build on the dataset by \cite{cerina2025possum}, which contains tweet–reply pairs, tweet metadata, and user-level information. Each data point includes a tweet, its parent tweet, and the identity of the replying user. We split the dataset into training and test sets and focused on $250$ users with at least $20$ replies in the test set. For each of these users, we randomly sampled $20$ reply tweets.

Similarly, we also used Reddit data from the \texttt{r/politics} subreddit, obtained from the Pushshift dataset~\citep{baumgartner2020pushshift}. Here, we selected 492 users and, as with the Twitter/X sample, randomly drew $20$ reply messages per user following a train–test split.

Finally, the Bluesky data is drawn from the BluePrint dataset~\citep{buck2025blueprint}, a resource for persona-based language modeling curated from the Bluesky platform.
Also from this dataset, we selected 59 user instances, each represented by $20$ reply messages after a train–test split. These instances correspond to clusters of similar accounts constructed by the dataset creators rather than to individual users. To make them compatible with the other datasets, we generated synthetic user identifiers using ChatGPT-4o.

\subsection{Models}
Our objective is to simulate user responses to tweets with LLMs and to systematically evaluate the extent to which AI-generated replies align with human responses in terms of style, semantics, and tone. To this end, we prompted LLMs to produce one-sentence responses to a real-world social media message, emulating each user's linguistic style and conversational behavior.

We evaluated nine open-weight LLMs from six distinct model families, namely \textbf{Apertus}, \textbf{DeepSeek}, \textbf{Gemma}, \textbf{Llama}, \textbf{Mistral}, and \textbf{Qwen}. Our analysis primarily focused on compact models in the $4$–$8$B parameter range, which are increasingly used in research and applications due to their efficiency and accessibility. For comparison, we also included a large-scale model, \textbf{Llama~3.1~70B}~\citep{meta_llama_3_1_70b}, to benchmark the performance gap across model scales. Specifically, we tested Apertus-8B-2509~\citep{swissai2025apertus}, DeepSeek-R1-Distill-Llama-8B~\citep{deepseekai2025deepseekr1incentivizingreasoningcapability}, Gemma~3~4B~Instruct~\citep{gemma_2025}, Llama~3.1~8B~\citep{meta_llama_3_1_8b}, Llama~3.1~8B~Instruct~\citep{meta_llama_3_1_8b_instruct}, Llama~3.1~70B~\citep{meta_llama_3_1_70b}, Mistral~7B~v0.1~\citep{jiang2023mistral7b}, Mistral~7B~Instruct~v0.2~\citep{jiang2023mistral7b}, and Qwen~2.5~7B~Instruct~\citep{qwen2}. Each model was run with a temperature of~$0.8$, using prompts that included a persona description generated by ChatGPT-4o based on a sample of~$100$ user replies from the training set.

\subsection{Computational Turing Test}
Our evaluation comprises both aggregate quality metrics and fine-grained interpretable analysis. We first assess overall generation quality through two complementary lenses: automated detectability and semantic preservation. We then conduct a deeper analysis to understand which specific linguistic characteristics distinguish AI-generated from human-authored content.

\subsubsection{Detectability: BERT-based Turing test}
We assess how effectively large language models generate social media replies that remain indistinguishable from human-authored content when evaluated by automated classifiers. We employ a BERT-based binary classification model to distinguish between the two text types. To construct the classification dataset, we collect all AI-generated replies and pair them with an equal number of human-written messages randomly sampled from the training set, creating a balanced dataset. We then train the BERT-based sequence classification model to discriminate between AI-generated and human-authored text, evaluating its performance on a held-out validation set. To ensure statistical robustness, we repeat this procedure three times using different random seeds and report the average accuracy across runs. Higher classification accuracy indicates more distinguishable AI-generated content, with accuracy approaching $50\%$ (random chance), suggesting successful human mimicry.

\subsubsection{Semantic Matching: Cosine Similarity score}
We examine the semantic similarity between AI-generated text and the corresponding human ground truth. We embed each reply into a dense vector space using the \texttt{all-MiniLM-L6-v2} model from the \texttt{SentenceTransformers} library. For each pair consisting of a human-authored reply and its corresponding AI-generated counterpart, we compute the cosine similarity between their embeddings. We then analyze the distribution of these cosine similarity scores across all reply pairs. This metric provides a complementary perspective to detectability, measuring how well AI-generated content preserves the semantic content of human responses.

\subsubsection{Interpretable Feature Analysis: Random Forest model and Empath library}
While the above metrics provide aggregate measures of distinguishability and similarity, they do not reveal which specific linguistic characteristics drive these differences. To understand what makes AI-generated content distinct from human-authored text, we employ two complementary approaches examining different dimensions of language.

First, we train a Random Forest classifier based on interpretable stylistic and structural features. The feature set includes a broad range of interpretable textual indicators capturing lexical, stylistic, affective, and structural dimensions of online conversation. Specifically, we compute:
\begin{itemize}
\item \textit{Lexical/structural features}: word count, character count, average word length, sentence length variance, punctuation count, uppercase count, mention count, hashtag count, link count, emoji count, transition word count, superlative count, and hedge word count (based on a predefined lexicon).
\item \textit{Diversity and repetitiveness}: type–token ratio (lexical diversity), 3-gram repetition count, and a proxy for perplexity (self-entropy of the word distribution).
\item \textit{Stylistic/affective features}: sentiment polarity scores (via \texttt{VADER}), toxicity score (using the \texttt{unitary/toxic-bert} model).
\item \textit{Syntactic complexity}: average clauses per sentence, approximated by the number of commas plus one.
\end{itemize}

Second, to examine topical and semantic patterns, we employ the Empath library~\cite{fast2016empath}. Empath identifies and quantifies the presence of $194$ built-in topical categories (e.g., emotions, social relationships, cognitive processes) by analyzing lexical patterns within the text. For each reply, we extract normalized topic signal scores across all categories. We then compare the distributions of these scores between AI-generated and human-authored samples using the Wilcoxon rank-sum test for each Empath category. To account for multiple comparisons across the 194 categories, we apply the Benjamini-Hochberg false discovery rate (FDR) correction and identify features as significantly different when the adjusted p-value is below 0.05.

\section{State-of-the-Art Results}
We begin by using our validation framework to evaluate the established state-of-the-art for calibrating LLMs to generate human-like messages.

\subsection{BERT-based Turing test}
\begin{figure}[h!]
    \centering
    \includegraphics[width=.5\columnwidth]{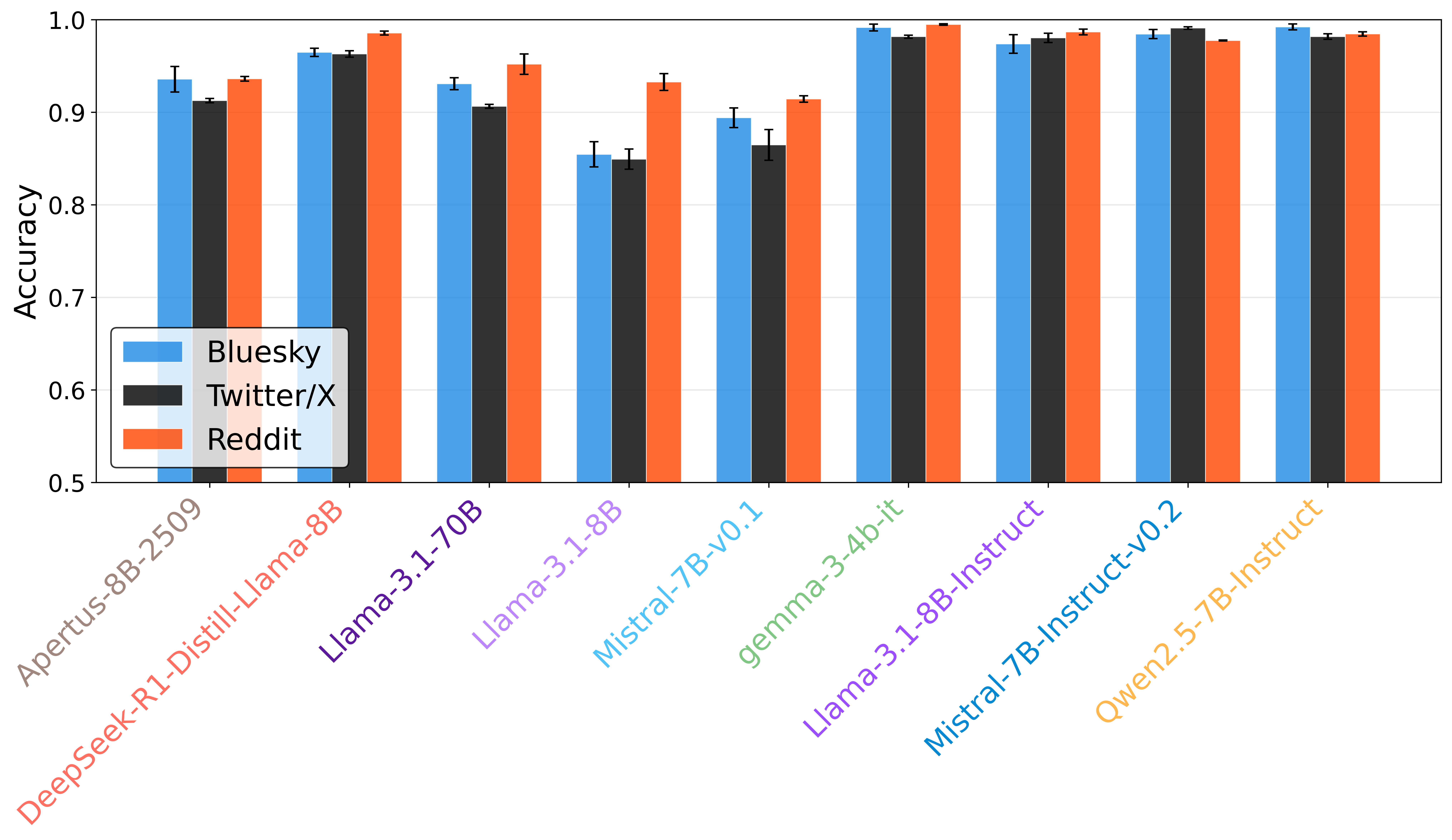}
    \caption{BERT classifier accuracy for optimal model configurations across datasets and optimization strategies. Lower accuracy indicates better human mimicry (approaching 50\% chance level). Non-instruction-tuned models from Llama, Mistral, and Apertus demonstrate superior deception capabilities compared to instruction-tuned variants. Twitter/X content (black bars) proves most susceptible to successful LLM replication across all three social media platforms examined.}
    \label{fig:baseline_accuracy}
\end{figure}

As shown in Figure~\ref{fig:baseline_accuracy}, the BERT classifier achieves consistently high accuracy across all models and datasets, never falling below $85\%$. This indicates that current state-of-the-art LLMs produce replies that remain readily distinguishable from human-authored content when evaluated by automated classifiers. Furthermore, we observe notable variation across model types, with non-instruction-tuned variants from Llama, Mistral, and Apertus achieving the lowest detection rates (i.e., best human mimicry). Platform-specific differences also emerge: the Twitter/X dataset consistently yields lower classification accuracy compared to Bluesky and Reddit, suggesting that LLMs more successfully replicate the conversational patterns of Twitter/X users. This may reflect either the distinctive stylistic characteristics of different platforms or their varying representation in LLM training corpora.

\subsection{Cosine Similarity}
\begin{figure}[h!]
\centering
    \includegraphics[width=.5\columnwidth]{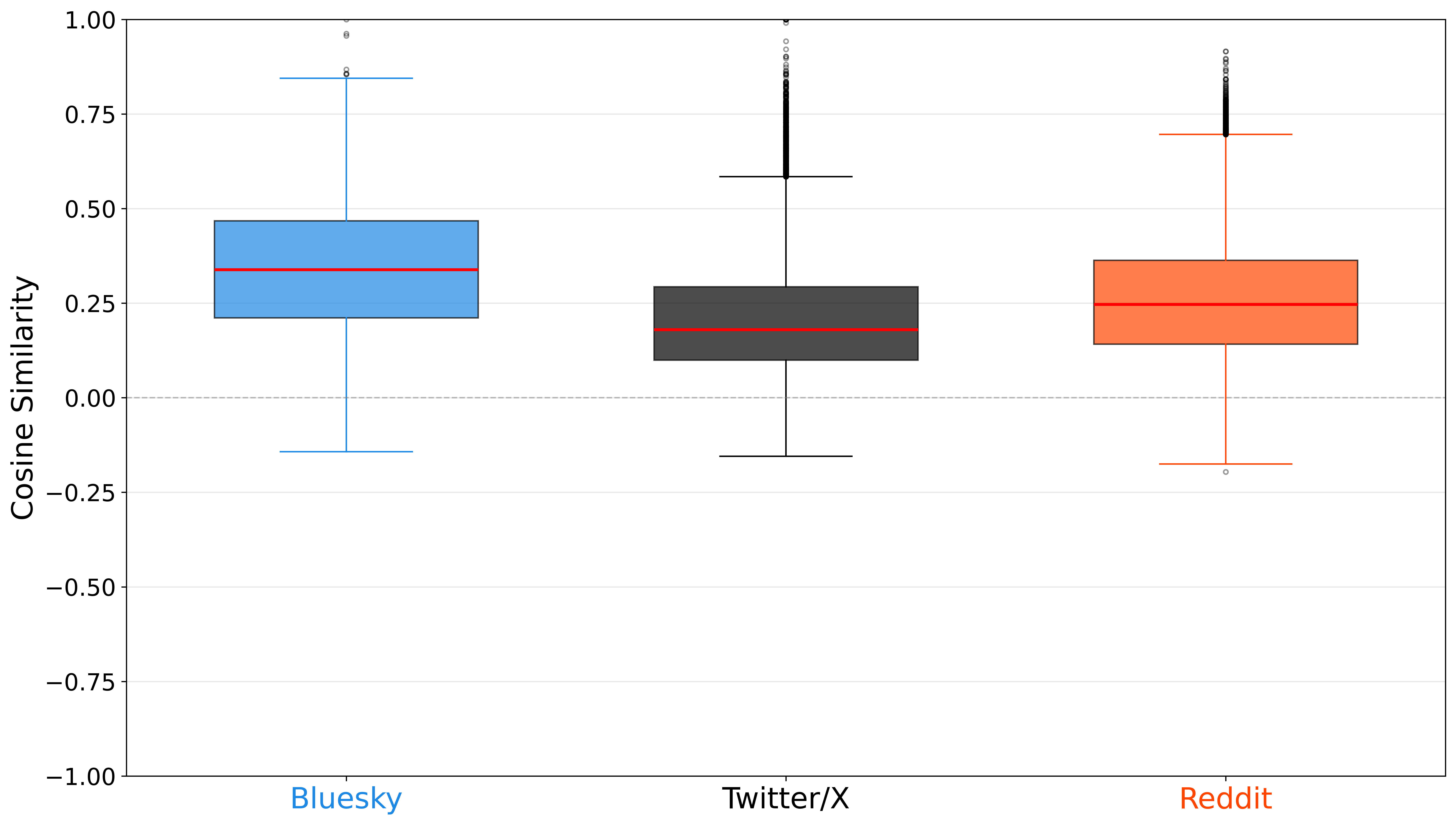}
    \caption{Distribution of content similarity between the AI-generated response and the ground-truth response, measured via the cosine similarity score using the \texttt{all-MiniLM-L6-v2} model from the \texttt{SentenceTransformers} library.}
    \label{fig:cosine_similiarity_baseline}
\end{figure}

With respect to semantic similarity scores, measured via cosine similarity of sentence embeddings, Figure~\ref{fig:cosine_similiarity_baseline} reveals modest alignment between AI-generated and human-authored replies. Median similarity scores range from approximately $0.2$ for Twitter/X to $0.3$ for Bluesky, with relatively consistent performance across different LLM architectures. While these scores indicate that AI-generated replies capture some semantic aspects of human responses, the moderate similarity values suggest substantial divergence in content. This limited semantic alignment likely reflects the challenging nature of the task: LLMs generate replies without access to the full conversational context available to human authors, requiring them to infer appropriate responses from limited information. The consistency across models further suggests that current architectural differences may be less important than the fundamental challenge of context-limited generation. Together, these metrics reveal that in generating LLM-powered social media replies significant gaps remain in both semantic fidelity and stylistic human-likeness, motivating our deeper interpretable analysis of the specific linguistic characteristics that distinguish AI-generated from human-authored content.

\subsection{Interpretability Results}
To understand which specific linguistic characteristics distinguish AI-generated from human-authored content, we examine both stylistic features through Random Forest classification and topical patterns through Empath analysis.

\subsubsection{Stylistic and Structural Features}
\begin{figure*}
    \centering
    \includegraphics[width=\linewidth]{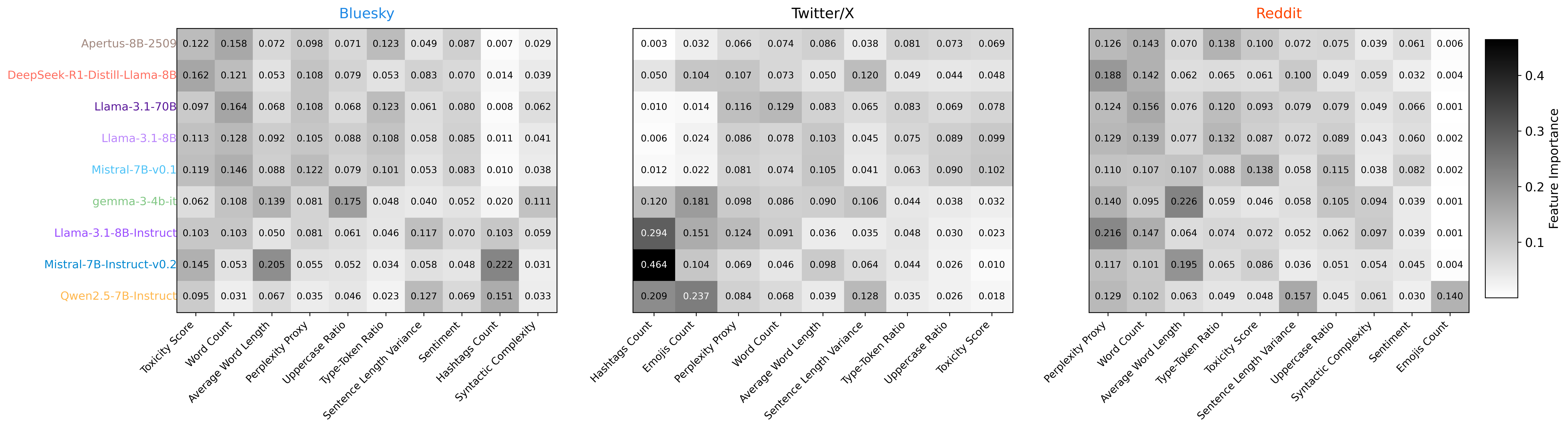}
    \caption{Results of the feature importance analysis derived from the Random Forest model trained to classify AI-generated vs Human text. The top $10$ most important features are displayed for each dataset and model. For each dataset, features are ordered left to right according to decreasing importance across all models.}
    \label{fig:ML_baseline}
\end{figure*}

The Random Forest classifier reveals both platform-specific and model-specific patterns in feature importance (Figure~\ref{fig:ML_baseline}). Across all three platforms, toxicity score consistently emerges as a critical discriminator, appearing among the top features for nearly all models. Beyond toxicity, platform-specific patterns emerge in the discriminative features. For Bluesky, word count, average word length, and perplexity proxy show relatively high importance across models, suggesting that AI-generated content differs in verbosity, as well as lexical sophistication and diversity. Twitter exhibits a distinct pattern: hashtag count, emoji count, and perplexity proxy emerge as strong discriminators. The prominence of platform-specific features (hashtags, emojis) suggests that AI struggles to replicate the distinctive stylistic conventions of Twitter discourse. Reddit classification shows smaller variability across models but consistently relies on perplexity proxy, word count, and average word length, pointing to differences in linguistic complexity and response elaboration.

Notably, the relative importance of specific features varies substantially across models within each platform, as evidenced by the varying shades across rows. Some models show pronounced reliance on particular features (darker cells), while others distribute importance more evenly. For instance, on Twitter, Llama-3.1-8B-Instruct and Mistral-7B-Instruct-v0.2 show particularly strong signals from hashtag count, while Qwen-2.5-7B-Instruct differs in reproducing human-like levels of emojis across all the datasets. This heterogeneity suggests that different LLM architectures fail to replicate human writing in distinct ways, with each model producing distinctive linguistic signatures that enable classification.

\subsubsection{Topical and Affective Patterns}
\begin{figure*}[t]
    \centering
    \includegraphics[width=\linewidth]{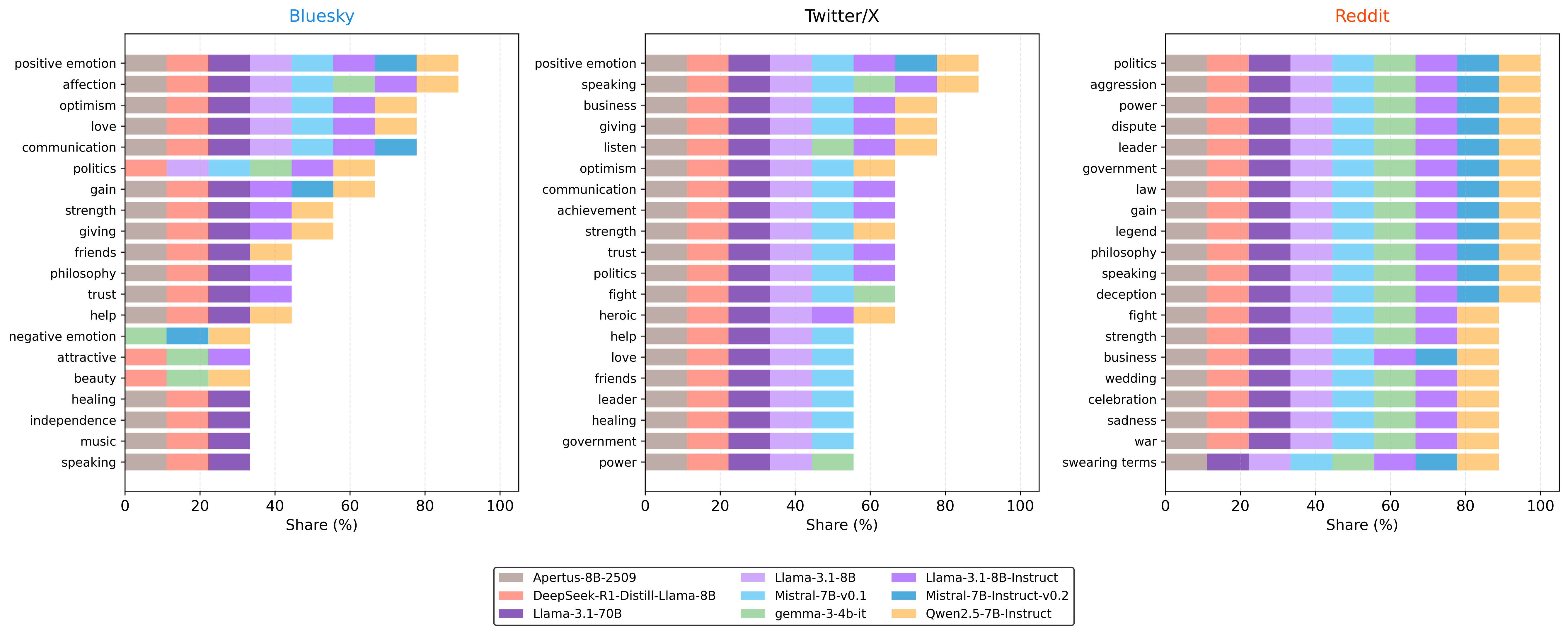}
    \caption{Empath library analysis~\cite{fast2016empath}. For each dataset, the plot shows up to $20$ most common (across all nine models) statistically significantly different features.}
    \label{fig:empath_baseline}
\end{figure*}

The Empath analysis reveals systematic divergences in topical and emotional content across all platforms (Figure~\ref{fig:empath_baseline}). All models exhibit substantial difficulties in replicating human topical patterns, though the severity varies by model and platform. In the Reddit dataset, all nine models differ significantly on at least 12 of the top 20 most frequently divergent features, with seven models showing significant differences on an additional eight features. The degree of divergence varies by platform: Bluesky generally shows the fewest significantly different features, followed by Twitter, with Reddit exhibiting the most pronounced topical gaps. Interestingly, not only highly detectable models such as DeepSeek-R1 but also some models that achieve lower BERT detectability scores, such as Llama-3.1-8B and Apertus, display among the highest number of significantly different Empath features, suggesting that evading classifier detection does not necessarily correspond to topical and affective alignment with human content.

Across platforms, affective and social categories emerge as consistent discriminators: positive emotion, affection, optimism, and communication appear prominently in the top differentiating features for multiple datasets. Platform-specific patterns also emerge. Twitter shows strong differences in categories like speaking, business, and giving, while Reddit diverges notably on politics, aggression, and power. Bluesky exhibits differences across emotional categories, including both positive and negative emotions. These patterns suggest that LLMs struggle to replicate the affective tone and social-relational language characteristic of human social media discourse, with the specific topical gaps varying according to platform norms and conversational culture.

Together, these interpretable analyses reveal that the detectability of AI-generated content stems from both topical-affective gaps (particularly in emotional expression and social language, as shown by Empath) and stylistic-structural differences (especially in toxicity, lexical characteristics, and platform-specific conventions, as shown by Random Forest). These patterns persist across different LLM architectures and platforms, though the specific discriminative features vary, suggesting fundamental limitations in current models' ability to capture the full range of human conversational behavior.

\section{Prompt-Optimization Strategies}

\subsection{From zero-shot prompting to fine-tuning}
The evaluation of the state-of-the-art method reveals persistent gaps between AI-generated and human-authored content across semantic, stylistic, and topical dimensions. To investigate whether these gaps can be mitigated through enhanced generation strategies, we systematically evaluate five prompt configurations of increasing complexity. Each configuration builds incrementally on the previous one, incorporating additional contextual information and model adaptation techniques designed to improve the fidelity of AI-generated replies to human writing patterns.

We test the following configurations:
\begin{enumerate}
\item \textbf{Baseline (BL)}: A simple prompt requesting the model to generate a concise reply to a given social media message, without any user-specific information or stylistic guidance.
\item \textbf{Persona (PE)}: The prompt is augmented with a user persona description, which is generated by GPT-4o based on the full set of the user's replies in the training set. This persona captures the user's interests, communication style, and typical topics of discussion. \textit{This configuration corresponds to the state-of-the-art method evaluated in the previous section.}
\item \textbf{Persona + Stylistic Examples (PE+SE)}: Building on the persona description, the prompt includes 10 examples of the user's prior replies drawn from the training set, providing concrete demonstrations of their writing style.
\item \textbf{Persona + Stylistic Examples + Context Retrieval (PE+SE+CR)}: The prompt is further augmented with user-specific contextual information retrieved from prior posts, using a similarity-based retrieval method similar to the approach proposed in~\cite{tan2024democratizing}. This provides relevant background information that may inform the reply.
\item \textbf{Persona + Stylistic Examples + Context Retrieval + Fine-tuning (PE+SE+CR+FT)}: The baseline model is fine-tuned on the full training set using the Parameter-Efficient Fine-Tuning (PEFT) library~\citep{peft}, adapting the model's parameters to the user's specific writing patterns. The fine-tuned model is then prompted with persona, examples, and retrieved context.
\end{enumerate}

We apply these configurations across all nine LLMs and three datasets, evaluating their impact on the metrics established in our baseline analysis: BERT classification accuracy (detectability), cosine similarity (semantic fidelity), Random Forest feature importance (stylistic alignment), and Empath feature divergence (topical alignment). By comparing the Persona configuration against both simpler (Baseline) and more sophisticated approaches (PE+SE, PE+SE+CR, PE+SE+CR+FT), this systematic evaluation allows us to identify which optimization strategies most effectively reduce the gap between AI-generated and human-authored social media content, and whether different platforms or models benefit differentially from specific approaches.

\begin{figure}
    \centering
    \includegraphics[width=\linewidth]{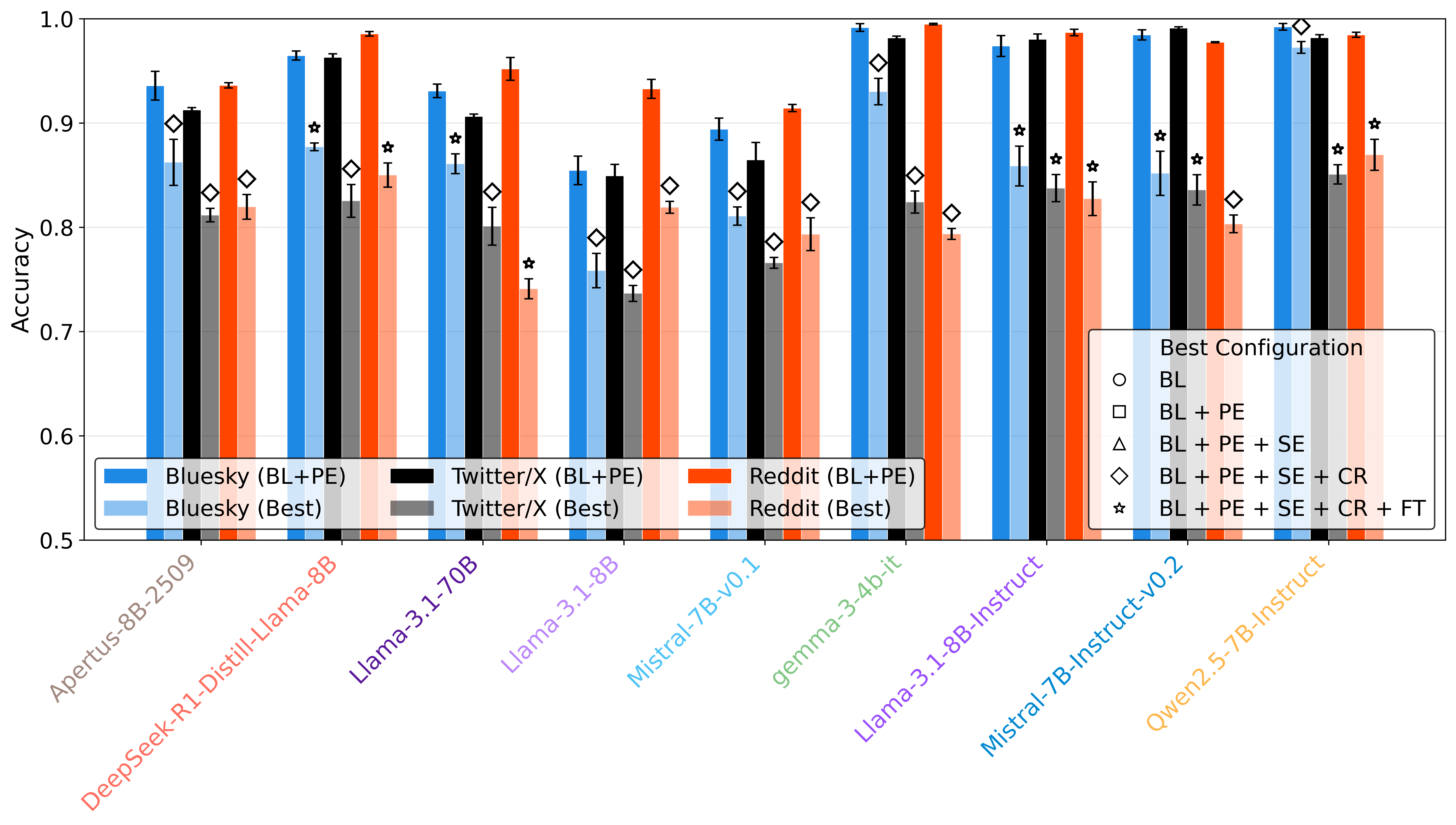}
    \caption{BERT classifier accuracy comparing state-of-the-art method (Baseline + Persona) and best-performing configurations across models and datasets. Increasing configuration complexity definitely improves the performance in making the models less detectable. Furthermore, it substantially reduces the differences between the different models. Also, the best performing configuration is either the one that integrates stylistic examples and context (besides the persona description), or the one that also uses a fine-tuned model (especially in the case of instruct-tuned models).}
    \label{fig:best_model_performance}
\end{figure}

\subsection{Accuracy improvement in BERT-based Turing test}

Figure~\ref{fig:best_model_performance} presents the BERT classification accuracy for the best-performing configuration of each model, compared against the state-of-the-art (Persona) configuration evaluated in the previous section. The results demonstrate that increasing configuration complexity substantially improves human-likeness: the best configurations achieve markedly lower classification accuracies than the baseline across all models and datasets (with the exception of Qwen-2.5-7B-Instruct on the Bluesky dataset), with improvements of roughly $10\%$ and peaks reaching $15-20\%$. This improvement is particularly pronounced for the previously low-performing instruction-tuned models, which benefit more from additional optimization steps, and in particular from fine-tuning. The absolute performance hierarchy observed in the baseline persists: non-instruction-tuned variants (Llama-3.1-8B, Llama-3.1-70B, Mistral-7B-v0.1, and Apertus-8B-2509) almost consistently achieve the lowest classification accuracies (with values ranging between $75$ and $85\%$). Notably, optimization reduces the variance across models: the best configurations produce more uniform performance, with differences between models diminishing substantially compared to the baseline.

Platform-specific patterns also remain consistent, with Twitter/X content exhibiting the lowest detectability.

\begin{figure}
    \centering
    \includegraphics[width=.7\linewidth]{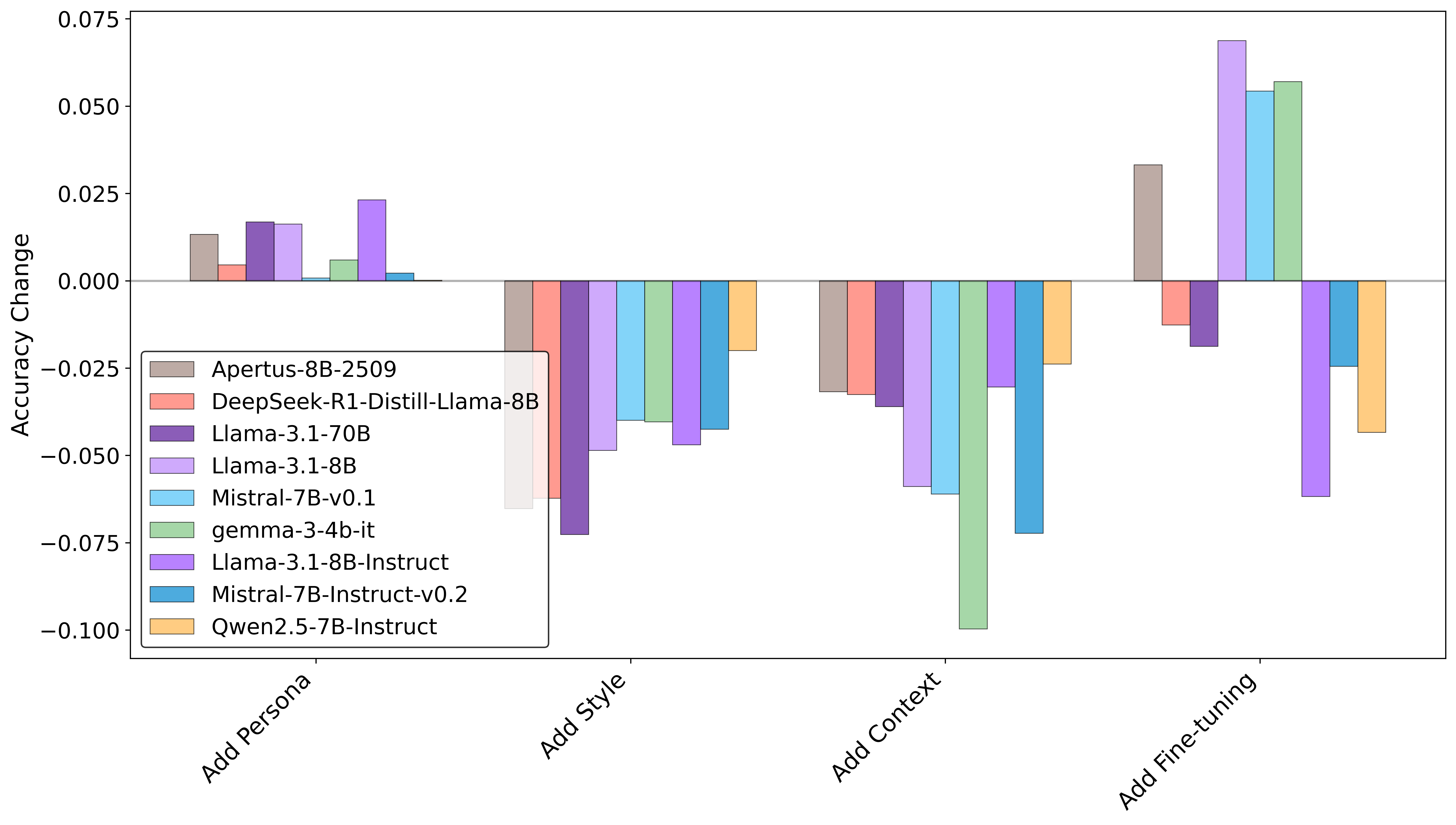}
    \caption{Stepwise impact of prompt-optimization strategies on BERT classifier accuracy, aggregated across datasets. Each point represents the change in accuracy when adding a new component: BL→PE (baseline to persona), PE→PE+SE (adding stylistic examples), PE+SE→PE+SE+CR (adding context retrieval), and PE+SE+CR→PE+SE+CR+FT (adding fine-tuning). Negative changes (downward) indicate reduced detectability and improved human mimicry. Stylistic examples and context retrieval consistently reduce classification accuracy, while persona descriptions and fine-tuning show mixed or adverse effects.}
    \label{fig:accuracy_improvement_per_step}
\end{figure}

A closer examination of the stepwise progression through configurations reveals a surprising pattern. Figure~\ref{fig:accuracy_improvement_per_step} illustrates how each incremental enhancement affects detection performance. Counterintuitively, not all optimization steps reduce detectability. The transition from baseline to persona (BL→PE) produces negligible or even slightly positive changes in accuracy for most models, indicating that adding abstract persona descriptions does not improve, and may even deteriorate, human-likeness. Similarly, fine-tuning (PE+SE+CR→PE+SE+CR+FT) shows highly variable effects, with some models experiencing substantial increases in detectability, particularly among non-instruction-tuned variants.

In stark contrast, two interventions consistently reduce classification accuracy across nearly all models: adding stylistic examples (PE→PE+SE) and incorporating context retrieval (PE+SE→PE+SE+CR). These steps produce the most uniform improvements, with stylistic examples yielding accuracy reductions of approximately $2-6\%$ and context retrieval contributing an additional $2-10\%$ across models. The consistency of these effects suggests that concrete demonstrations of writing style and relevant contextual information address fundamental gaps in LLM-generated content that abstract descriptions or parameter adaptation cannot overcome.

These findings challenge conventional assumptions about what makes AI-generated text more human-like. While persona descriptions and fine-tuning might seem like sophisticated optimization strategies, they appear less effective, and sometimes counterproductive, compared to providing concrete stylistic examples and relevant context. This pattern suggests that current LLMs struggle less with adapting their general capabilities to specific users and more with accessing the specific stylistic and contextual cues that inform natural human responses.

\subsection{Semantic Similarity Declines Despite Reduced Detectability}

\begin{figure}[tbp!]
\centering
\includegraphics[width=.5\columnwidth]{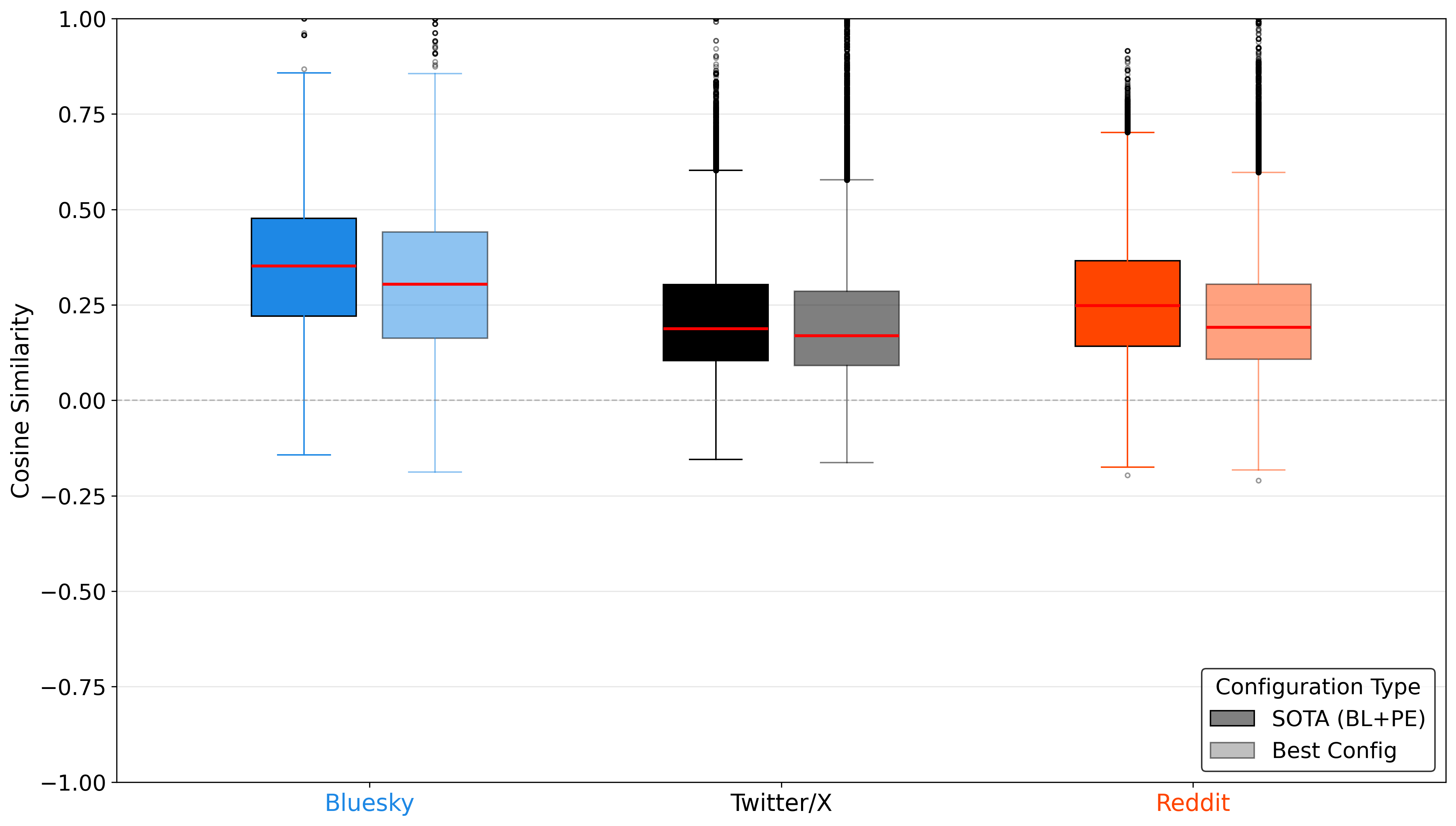}
\caption{Distribution of cosine similarity scores between AI-generated and ground-truth human replies, comparing state-of-the-art (SOTA: BL+PE) and best-performing configurations across datasets. Counterintuitively, configurations that achieve lower BERT detectability show decreased semantic similarity to human ground-truth. Median scores decline by 0.07 for Bluesky (0.34→0.26), 0.01 for Twitter/X (0.18→0.16), and 0.04 for Reddit (0.25→0.20), suggesting a trade-off between stylistic human-likeness and semantic fidelity. Box plots show median (red line), interquartile range (box), and full distribution excluding outliers (whiskers).}
\label{fig:cosine_similiarity_optimal_config}
\end{figure}

While the optimization strategies substantially reduce BERT classification accuracy, they surprisingly lead to decreased semantic similarity to ground-truth human replies. Figure~\ref{fig:cosine_similiarity_optimal_config} compares the cosine similarity distributions between the state-of-the-art (Baseline + Persona) configuration and the best-performing configurations identified in the previous section. Across all three platforms, the best configurations show lower median cosine similarity scores than the baseline, revealing an unexpected trade-off. For Bluesky, the median similarity decreases from 0.34 (SOTA) to 0.28 (Best Config). Twitter/X shows a more modest but consistent decline from 0.18 to 0.16, and Reddit exhibits an intermediate drop from 0.25 to 0.20.

This counterintuitive finding reveals a fundamental tension in AI-generated content: optimizing for human-likeness as judged by classifiers can come at the expense of semantic fidelity to what humans actually say. The best-performing configurations, which incorporate stylistic examples and contextual information, appear to prioritize matching the general stylistic and affective characteristics of human social media discourse over reproducing the specific semantic content of individual reference replies. In other words, these models learn to write in a more human-like manner but not necessarily to say the same things humans would say in response to specific prompts.

This dissociation suggests that detectability and semantic alignment represent fundamentally different dimensions of generation quality. Configurations that successfully evade classifiers do so by capturing the distributional properties of human writing style, tone, and affect, but this stylistic mimicry may actually obscure or dilute the semantic signal from the original context. The decline in cosine similarity indicates that as models become better at ``sounding human'', they may drift further from contextually appropriate responses, trading semantic relevance for stylistic authenticity.

\subsection{Feature-Level Changes Under Optimization}

To understand how prompt optimization alters the linguistic characteristics that distinguish AI-generated from human-authored content, we apply the same interpretable analysis framework to the best-performing configurations.

\begin{figure*}[h]
    \centering
    \includegraphics[width=\linewidth]{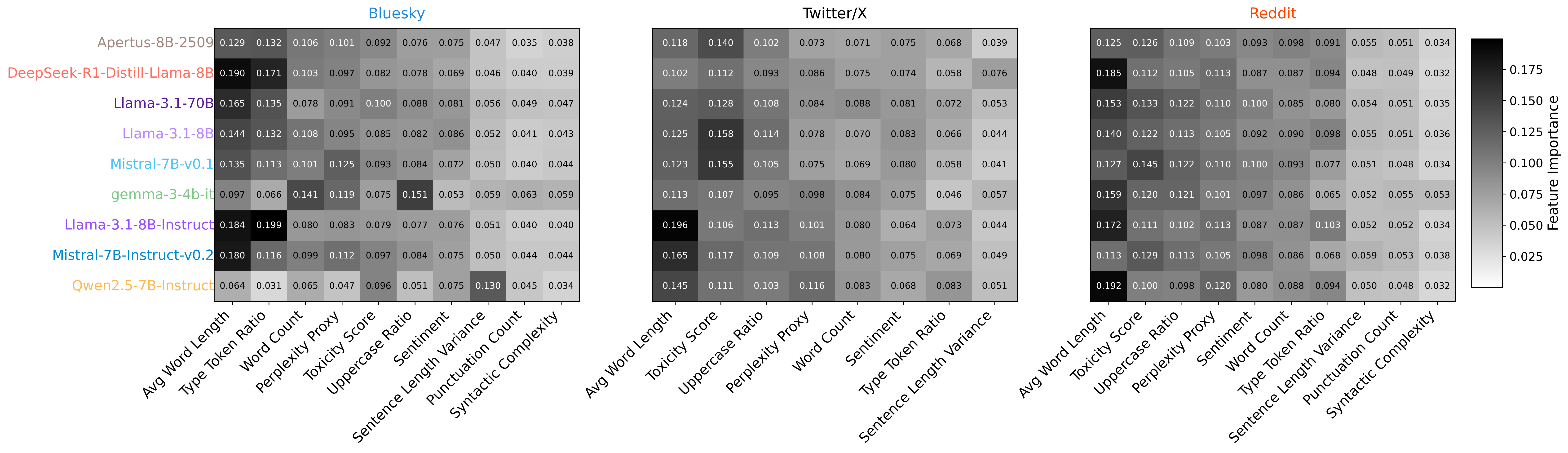}
    \caption{Random Forest feature importance analysis for best-performing configurations across models and datasets. The top 10 most important features are displayed for each dataset and model, ordered left to right by decreasing average importance across all models. Darker cells indicate higher discriminative power. Comparison with baseline results (Figure~\ref{fig:ML_baseline}) reveals a dramatic shift: average word length emerges as the dominant discriminator across all platforms, while platform-specific formatting features (hashtags, emojis) diminish in importance. Toxicity score persists and even increases in relative importance for Twitter and Reddit, indicating that affective divergences remain despite optimization.}
    \label{fig:ML_config_optmization}
\end{figure*}

The Random Forest feature importance analysis (Figure~\ref{fig:ML_config_optmization}) reveals significant shifts in which features distinguish AI from human text compared to the baseline (Figure~\ref{fig:ML_baseline}). Most notably, average word length emerges as the dominant discriminator across all three platforms in optimized configurations, whereas it played a more modest role in the baseline. This suggests that optimization strategies, while reducing overall detectability, amplify (relatively to the other features) systematic differences in lexical sophistication.

Platform-specific patterns also shift substantially. For Bluesky, the baseline's reliance on toxicity score, word count, and average word length transforms into a focus on average word length, type-token ratio, and word count in optimized configurations, with toxicity becoming less prominent. Twitter shows the most dramatic reorganization: the baseline emphasized hashtag count, emoji count, and perplexity proxy, while optimized configurations prioritize average word length, toxicity score, and uppercase ratio. This shift suggests that optimization successfully addresses platform-specific formatting conventions (hashtags, emojis) but places higher importance on linguistic complexity and affective intensity. Reddit maintains some consistency, with perplexity proxy remaining important, but average word length and toxicity score gain prominence in optimized configurations.

Critically, toxicity score persists as a discriminator across platforms, though its relative position varies. For Twitter and Reddit, toxicity actually increases in importance under optimization, appearing among the top three features. This indicates that optimization strategies do not address affective divergences as effectively as other features.

\begin{figure*}[h]
    \centering
    \includegraphics[width=\linewidth]{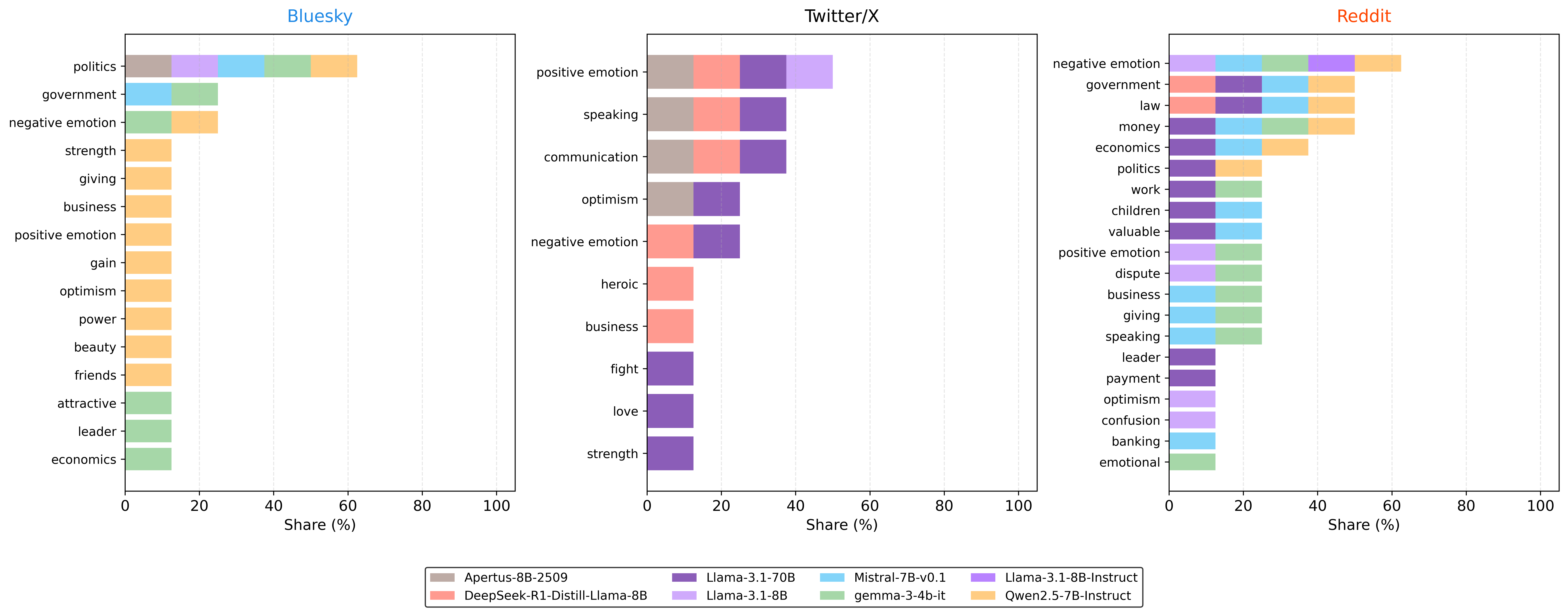}
    \caption{Empath topical divergence analysis for best-performing configurations~\cite{fast2016empath}. For each dataset, the plot shows up to 20 Empath categories that most frequently exhibit statistically significant differences between AI-generated and human-authored text across all nine models. Each colored segment indicates one model's contribution. Compared to baseline (Figure~\ref{fig:empath_baseline}), the number of divergent features decreases dramatically, particularly for Twitter, where many models show minimal topical differences. Emotional categories (``positive emotion,'' ``negative emotion'') persist where differences remain, but the overall scope of topical divergence narrows substantially.}
    \label{fig:empath_optimal_config}
\end{figure*}

The Empath analysis reveals a striking reduction in topical-affective divergences under optimization, though certain emotional categories persist. Comparing Figure~\ref{fig:empath_optimal_config} to the baseline (Figure~\ref{fig:empath_baseline}), the number of statistically significantly different features decreases drastically across all platforms. For Twitter, the change is most dramatic: the baseline showed approximately 15-20 frequently divergent features across models, while the optimized configuration reduces this to only 7 features and 3 models, namely DeepSeek, Llama-3.1-8B, and Apertus. Bluesky shows a substantial reduction from approximately 20 features to about 15 (with only 5 models contributing), while Reddit maintains more divergent features but still shows meaningful improvement.

Despite this overall reduction, emotional categories persist among the top discriminators where differences remain: ``positive emotion'' continues to appear for Bluesky and Twitter, while ``negative emotion'' remains prominent for Reddit. However, the scope of divergence narrows considerably. Many affective and social categories that showed widespread differences in the baseline (``affection,'' ``optimism,'' ``communication,'' ``speaking'') either disappear entirely or show differences for only 1-2 models in optimized configurations. This indicates that optimization strategies successfully address many topical dimensions while a core set of emotional categories remains challenging. The dramatic reduction in divergent features, particularly for Twitter, where some models achieve near-perfect topical alignment, demonstrates that prompt engineering (and sometimes fine-tuning) can substantially improve topical-affective resemblance, even if they cannot fully eliminate all emotional divergences.

\subsection{Post-Generation Optimization}
While the previous sections focus on improving generation quality through prompt engineering and fine-tuning, we now investigate whether post-generation selection from multiple candidates can further reduce detectability. Following best-of-N sampling approaches \citep{manning2025general}, we let each model produce $20$ replies per test message. Among them, we then select an optimal candidate using two distinct optimization procedures. This multi-candidate approach offers key advantages: it enables exploration of natural variation in model outputs and allows us to identify which linguistic features can be mitigated through selection versus those that represent fundamental model limitations.

We define two selection procedures optimizing for different dimensions of human-likeness:

\textbf{Cosine Similarity Selection.} The first procedure selects the candidate reply with the highest semantic similarity to the ground-truth human reply. Each candidate is embedded into a dense vector space using the \texttt{all-MiniLM-L6-v2} model from the \texttt{SentenceTransformers} library, and we compute the cosine similarity between each candidate and the reference reply. The reply with the highest similarity score is selected. This method directly optimizes for semantic alignment with the actual human response.

\textbf{ML-Optimal Selection.} The second procedure employs a machine learning classifier to identify the candidate reply most likely to be misclassified as human-authored. Importantly, this classifier uses the same interpretable feature set described in our earlier Random Forest analysis (lexical/structural features, diversity metrics, stylistic/affective features, and syntactic complexity). The set of LLM-generated replies is partitioned into folds according to users. For each fold, we train a Random Forest classifier (100 estimators, fixed random seed) on the remaining folds using both LLM-generated replies and human-authored replies from the training data. The classifier is then applied to the held-out fold to estimate each candidate's likelihood of being misclassified as human. The reply with the highest misclassification probability is selected.

The two optimization procedures capture complementary aspects of human-likeness: cosine similarity emphasizes semantic alignment with the actual human response, while the ML-optimal selection emphasizes stylistic and affective resemblance to human writing patterns in general. Because the cosine method relies directly on the reference reply while the ML-optimal selection does not, the two approaches are not strictly comparable but provide different perspectives on achievable improvements through post-generation selection.

To understand the relationship between these selection methods and the variability in generated outputs, we examine the overlap between replies selected by each method. The overlap rate between ML-optimal and cosine similarity selection averages approximately 5-10\% across models and datasets, barely exceeding the 5\% random chance baseline (1 in 20 candidates). This minimal overlap confirms that the two methods optimize for fundamentally different characteristics and select distinct replies.

\subsection{Impact of Post-Generation Selection Strategies}

\begin{figure*}[t!]
    \centering
    \includegraphics[width=\linewidth]{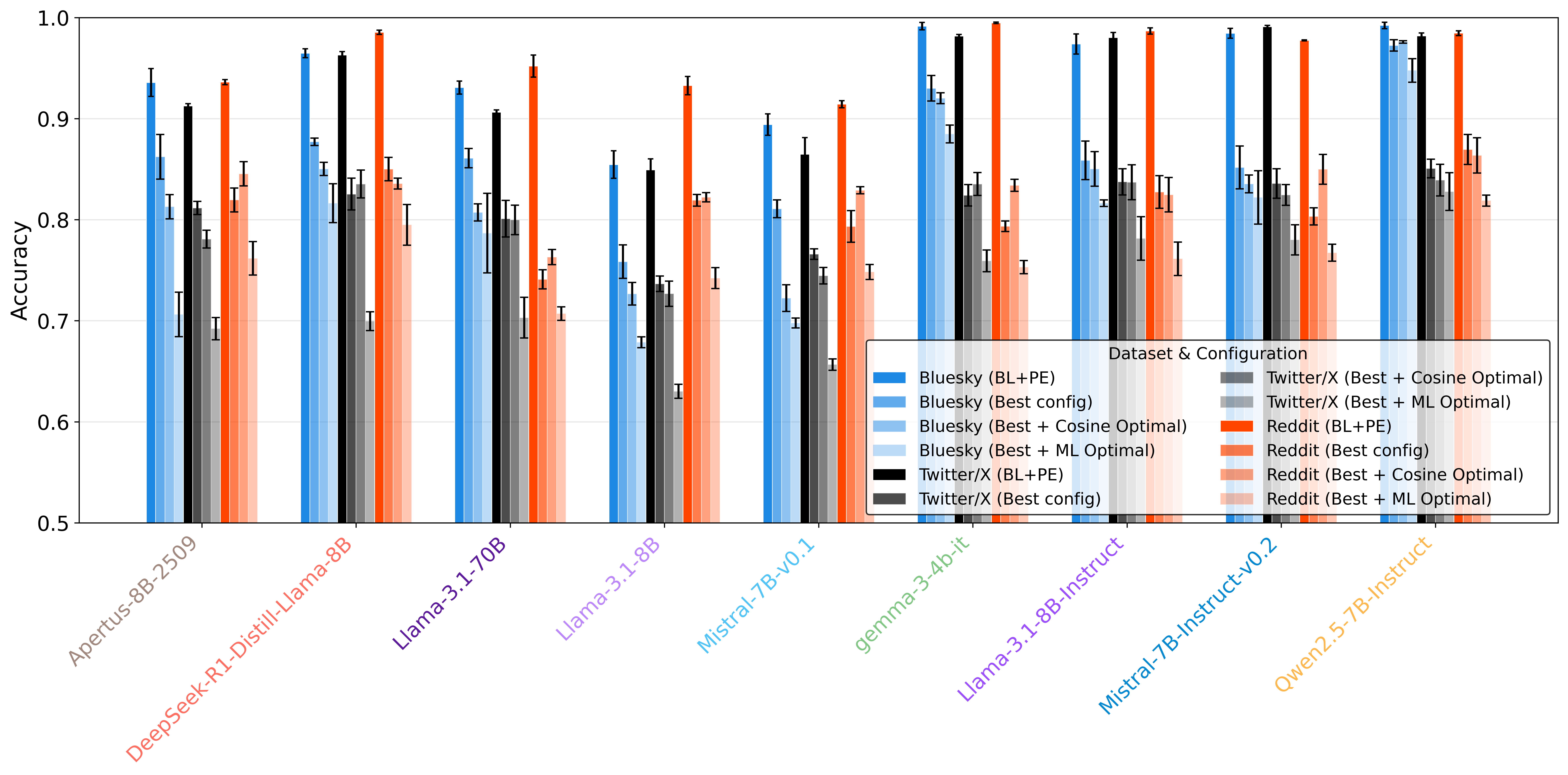}
    \caption{BERT classifier accuracy comparing post-generation optimization strategies across models and datasets. Four conditions are shown for each model: baseline configuration (BL+PE, darkest bars), best prompt-optimized configuration (Best config), and two post-generation selection methods applied to the best configuration (Cosine Optimal: selecting for highest semantic similarity to reference reply; ML Optimal: selecting for highest misclassification probability). Error bars represent standard deviation across three random seeds. Lower accuracy indicates better human mimicry. ML-optimal selection consistently achieves the lowest detectability, while cosine similarity selection shows variable effectiveness, often performing similarly to unselected best configurations.}
    \label{fig:post_generation_optimization}
\end{figure*}

Figure~\ref{fig:post_generation_optimization} compares the effectiveness of different post-generation selection strategies applied to the best-performing configurations. For each model and dataset, we evaluate four conditions: the baseline configuration (BL+PE), the best prompt-optimized configuration identified previously (Best config), and two post-generation selection methods applied to the best configuration (cosine similarity selection and ML-optimal selection).

Post-generation selection yields modest but measurable improvements in reducing detectability. The ML-optimal selection consistently achieves the lowest BERT classification accuracies across most models and datasets, typically reducing accuracy by 5-10 percentage points (with peaks of more than 15 for Apertus) compared to the best configuration without selection. Cosine similarity selection shows more variable performance: while it sometimes approaches ML-optimal effectiveness, it often performs comparably to or only marginally better than the best configuration alone. This asymmetry reflects the different optimization targets: ML-optimal selection directly optimizes for fooling a classifier with interpretable features correlated with BERT detectability, while cosine similarity optimizes for semantic alignment with the reference reply, which may not always correspond to reduced detectability.

Model-specific patterns reveal interesting variations. For some models (particularly Llama-3.1-8B, Mistral-7B-v0.1, and Apertus-8B-2509), ML-optimal selection produces substantial improvements, bringing accuracy closer to but never reaching the 50\% chance level. For others, such as Qwen-2.5-7B-Instruct, the gains from post-generation selection are often minimal, suggesting these models already generate relatively consistent outputs with limited exploitable variation across the 20 candidates.

\subsubsection{Semantic Similarity Under Post-Generation Selection}
\begin{figure}[h]
    \centering
    \includegraphics[width=\linewidth]{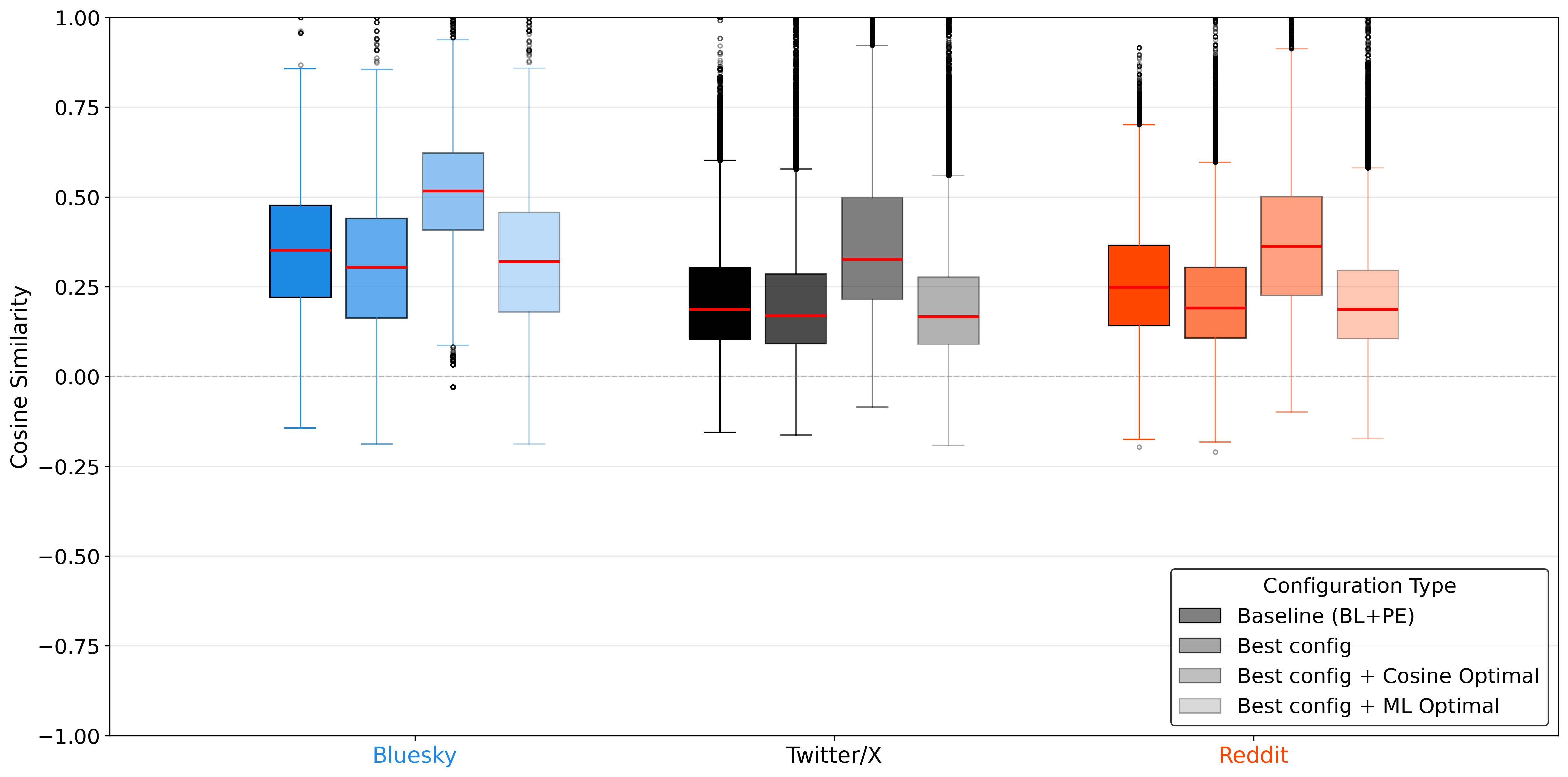}
    \caption{Distribution of cosine similarity scores between AI-generated and ground-truth human replies across all models, comparing four configuration types for each dataset: baseline (BL+PE, darkest), best prompt-optimized configuration without selection (Best config), best config with cosine-optimal selection (Best config + Cosine Optimal), and best config with ML-optimal selection (Best config + ML Optimal). Cosine-optimal selection substantially increases semantic similarity (medians: 0.51 Bluesky, 0.32 Twitter/X, 0.36 Reddit) compared to baseline (0.36, 0.19, 0.25) and other methods. ML-optimal selection produces the lowest similarity (0.32, 0.17, 0.19), demonstrating a trade-off between semantic fidelity and reduced detectability.}
    \label{fig:cosine_similarity_all_methods}
\end{figure}

The relationship between post-generation selection and semantic similarity reveals unexpected patterns that further illuminate the trade-offs between different optimization objectives (Figure~\ref{fig:cosine_similarity_all_methods}). Most strikingly, cosine-optimal selection substantially increases semantic similarity compared to both the baseline and unselected best configurations. For Bluesky, median similarity rises from 0.36 (baseline) to 0.51 (cosine-optimal), representing a 42\% improvement. Twitter/X shows median similarity increasing from 0.19 (baseline) to 0.32 (cosine-optimal), a 68\% gain. Reddit exhibits a similar pattern: median similarity climbs from 0.25 (baseline) to 0.36 (cosine-optimal), a 44\% increase. This demonstrates that when explicitly optimizing for semantic alignment by selecting from 20 candidates, models can produce replies substantially more similar to human ground truth than either baseline or best prompt-optimized configurations.

In contrast, the ML-optimal selection shows reduced semantic similarity compared to the baseline, similar to the results obtained without post-generation (just using the best model configuration). These values remain substantially below cosine-optimal levels, confirming that optimizing for reduced detectability moves generated content away from the semantic content of human replies.

The stark contrast between cosine-optimal and ML-optimal selection reveals a fundamental tension: the candidate pool of $20$ generated replies contains options that are either semantically similar to the reference reply or stylistically optimized to evade detection, but rarely both. When we select for semantic similarity, we can substantially exceed baseline performance, achieving median cosine scores of 0.32-0.51 across platforms. However, this semantic fidelity comes at the cost of increased detectability, as evidenced by the earlier finding that cosine-optimal selection does not consistently reduce BERT classification accuracy. Conversely, selecting for reduced detectability (ML-optimal) sacrifices semantic alignment, producing replies that sound human-like but say different things than the actual human responses.

This dissociation suggests that current optimization strategies, while increasing the diversity of generated candidates, do not enable simultaneous optimization of semantic fidelity and stylistic human-likeness. The candidate space explored by optimized configurations appears to trade off these dimensions: generating more human-like style (which ML-optimal selection exploits) or more semantically aligned content (which cosine-optimal selection exploits), but not both within the same candidate. The ability of cosine-optimal selection to substantially exceed baseline semantic similarity demonstrates that models possess the capacity for greater semantic alignment, but this capacity remains orthogonal to the stylistic characteristics that reduce detectability.

\section{Discussion}
This study provides the first systematic evidence of how far large language models (LLMs) have come -- and how far they still have to go -- in reproducing human communication. Using social media as a real-world benchmark, we combined a scalable \emph{computational Turing test} with a comprehensive calibration benchmark to assess both the realism and the tunability of LLM-generated text. Across three platforms, nine open-weight models, and multiple optimization strategies, we found that even the most advanced systems remain reliably distinguishable from human users.

Despite substantial optimization, AI-generated social media text continues to diverge in consistent ways from human writing. Classifiers detect AI replies with 70–80\% accuracy -- far above chance -- indicating that true human-likeness remains elusive. Non-instruction-tuned models such as Llama-3.1-8B, Mistral-7B, and Apertus-8B outperform their instruction-tuned counterparts, suggesting that alignment training introduces stylistic regularities that make text more, rather than less, machine-like. Interestingly, scaling up model size does not improve human-likeness: the large Llama-3.1-70B performs on par with, or even below, smaller models. This suggests that scaling does not translate into more authentically human communication. Platform-specific variation further underscores that LLMs reproduce the linguistic ``surface'' of human communication more easily than its contextual richness: imitation is strongest on X, weaker on Bluesky, and weakest on Reddit, where conversational norms are more diverse.

Affective language remains the clearest marker of artificiality. While optimization reduces structural divergences such as sentence length or word count, differences in emotional tone persist. Both toxicity and sentiment consistently emerge as strong discriminators, and topical analyses confirm that emotional expression -- positive or negative -- remains the most resistant dimension of mimicry. These results suggest that while LLMs can reproduce the \emph{form} of online dialogue, they struggle to capture its \emph{feeling}: the spontaneous, affect-laden expression characteristic of human interaction.

Our comprehensive calibration tests challenge the assumption that more sophisticated optimization necessarily yields more human-like output. Persona descriptions and fine-tuning -- strategies often thought to personalize or contextualize LLMs -- produce negligible or even adverse effects on realism. In contrast, simple techniques such as providing stylistic examples or retrieving relevant context consistently reduce detectability.

This non-monotonic pattern reveals that greater complexity does not automatically translate to greater realism: optimization may help models adhere to platform norms but can simultaneously introduce new linguistic artifacts. For example, after optimization, average word length becomes the dominant feature distinguishing AI text from human text, replacing earlier platform-specific cues like hashtags or emoji usage. Rather than eliminating the ``signature'' of AI authorship, optimization redistributes it across new dimensions of language.

Perhaps our most important finding is the trade-off between semantic fidelity and stylistic human-likeness. Configurations that make LLM text harder to detect simultaneously reduce its semantic similarity to human reference replies, while those that preserve meaning become easier to classify as artificial. This tension holds across all platforms and optimization levels: ML-optimal selection minimizes detectability but sacrifices content accuracy, whereas cosine-optimal selection maximizes semantic alignment but not realism.

This dissociation suggests that LLMs face a structural dilemma. The features that make text \emph{sound} human -- stylistic variability, emotional nuance, contextual looseness -- are not the same features that make it \emph{mean} what humans mean. Efforts to optimize both dimensions simultaneously encounter diminishing returns, indicating that human-likeness and semantic accuracy are not aligned objectives but competing goals in current model architectures. This finding extends prior work on competing objectives in LLM optimization \citep{manning2025general}, revealing that the trade-offs between different quality dimensions persist even when using post-generation selection strategies.

These findings carry implications for both scientific and applied uses of LLMs. 
{Our results help explain an apparent paradox: while LLMs can pass human-judgment-based Turing tests, successfully deceiving casual observers~\citep{bouleimen2025collective}, our computational analysis reveals they remain systematically distinguishable from human text. Human evaluators attend to surface-level plausibility while missing deeper divergences in affective expression, semantic patterns, and linguistic structure.}
For tasks requiring accurate reproduction of meaning -- such as personalized communication or policy simulations -- the observed semantic gaps caution against treating LLMs as faithful proxies for human respondents. For applications prioritizing realism -- such as synthetic data generation or behavioral modeling—our results identify effective but limited strategies: combining non-instruction-tuned models with stylistic examples, context retrieval, and classifier-based post-selection can reduce detectability, but none achieve genuine indistinguishability.

The persistent role of affective features highlights a key challenge for future development: calibrating models not only in content but also in emotion. Addressing this may require new objectives that explicitly model affective expression or emotional coherence, rather than relying on stylistic imitation alone. At the same time, our framework offers practical insights for detection and regulation: while synthetic content remains detectable, the gap is narrowing, underscoring the need for continuous monitoring of online discourse as LLM quality improves.

More broadly, this work demonstrates the value of combining scalable quantitative metrics with interpretable linguistic analysis. The computational Turing test provides a scalable and reproducible benchmark for validation, while feature-level interpretation reveals where and why models diverge from human language. Together, these tools move the evaluation of human-like AI from anecdotal and subjective judgments toward systematic measurement. The persistent linguistic boundary we uncover -- between meaning and style, accuracy and affect -- marks not a failure of current technology, but a guidepost for understanding what "human communication" truly entails and how far artificial systems remain from achieving it.

\section*{Acknowledgements}
This research was supported by the Swiss National Science Foundation (SNSF) under grant number [IZTAZ1\_223462], as part of the DemDialogueAI project funded through the Trans-Atlantic Platform's Democracy, Governance, and Trust initiative. For more information, see \url{https://www.transatlanticplatform.com/demdialogueai/}.

\section*{Ethics Statement}
This study uses social media data from three platforms. Twitter/X data were originally collected by \cite{cerina2025possum} via the official Twitter/X API. Reddit data from the r/politics subreddit were obtained from the Pushshift dataset~\citep{baumgartner2020pushshift}. Bluesky data were drawn from the BluePrint dataset, a resource for persona-based language modeling curated from the Bluesky platform. All datasets consist exclusively of publicly available posts and associated metadata that do not include personal or sensitive information. In accordance with institutional guidelines and applicable legal frameworks, including data protection regulations, this research does not constitute human subjects research and therefore did not require ethical review. Nonetheless, we took care to handle the data in ways that respect user privacy and minimize potential risks of harm.

\section*{Data and Code Availability}
All code used for the analysis is publicly available on GitHub at \url{https://github.com/paganick/validation_demdia}. This repository includes the computational Turing test implementation, calibration strategies, and all analysis scripts used to generate the results reported in this paper.

\bibliographystyle{plainnat}
\bibliography{bibliography}

\end{document}